\documentclass[conference]{IEEEtran}
\IEEEoverridecommandlockouts
\usepackage{cite}
\usepackage{amsmath,amssymb,amsfonts}
\usepackage{algorithmic}
\usepackage{graphicx}
\usepackage{textcomp}
\usepackage{array}
\usepackage{comment}
\usepackage{multirow}
\usepackage{longtable}
\usepackage{rotating}
\usepackage{booktabs}
\usepackage{url}
\def\BibTeX{{\rm B\kern-.05em{\sc i\kern-.025em b}\kern-.08em
    T\kern-.1667em\lower.7ex\hbox{E}\kern-.125emX}}
    
\usepackage[dvipsnames]{xcolor}

\begin{document}

\title{LiDAR point-cloud processing based on\\ projection methods: a comparison} 

\author{Guidong Yang$^{1}$, Simone Mentasti$^{2}$, Mattia Bersani$^{3}$,\\ Yafei Wang$^{1}$, Francesco Braghin$^{3}$, Federico Cheli$^{3}$ 
\thanks{Supported by project TEINVEIN: TEcnologie INnovative per i VEicoli Intelligenti, CUP (Codice Unico Progetto - Unique Project Code): E96D17000110009 - Call ``Accordi per la Ricerca e l'Innovazione", cofunded by POR FESR 2014-2020 (Programma Operativo Regionale, Fondo Europeo di Sviluppo Regionale – Regional Operational Programme, European Regional Development Fund).}
\thanks{$^{1}$ School of Mechanical Engineering, Shanghai Jiao Tong University, 800 Dongchuan Road, Shanghai, China, {\tt\small wyfjlu@sjtu.edu.cn}}
\thanks{$^{2}$ Department of Electronics Information and Bioengineering, Politecnico di MIlano, p.zza Leonardo da Vinci 32, Milan, Italy, {\tt\small name.surname@polimi.it}}
\thanks{$^{3}$ Department of Mechanical Engineering, Politecnico di Milano, via La Masa 1, Milan, Italy, {\tt\small name.surname@polimi.it}}%
}

\maketitle

\begin{abstract}
An accurate and rapid-response perception system is fundamental for autonomous vehicles to operate safely. 3D object detection methods handle point clouds given by LiDAR sensors to provide accurate depth and position information for each detection, together with its dimensions and classification. The information is then used to track vehicles and other obstacles in the surroundings of the autonomous vehicle, and also to feed control units that guarantee collision avoidance and motion planning. Nowadays, object detection systems can be divided into two main categories. The first ones are the geometric based, which retrieve the obstacles using geometric and morphological operations on the 3D points. The seconds are the deep learning-based, which process the 3D points, or an elaboration of the 3D point-cloud, with deep learning techniques to retrieve a set of obstacles. This paper presents a comparison between those two approaches, presenting one implementation of each class on a real autonomous vehicle.
Accuracy of the estimates of the algorithms has been evaluated with experimental tests carried in the Monza ENI circuit. The position of the ego vehicle and the obstacle is given by GPS sensors with RTK correction, which guarantees an accurate ground truth for the comparison.
Both algorithms have been implemented on ROS and run on a consumer laptop.

\end{abstract}
 
\begin{IEEEkeywords}
3D object detection, fully convolutional neural network

\end{IEEEkeywords}

\section{Introduction}
Object detection and tracking allow the control unit of an autonomous vehicle to plan the proper driving actions to be taken, accounting for the surrounding environment. As a basic module of object tracking, object detection can be conducted in traditional handcrafted feature-based methods as well as deep learning methods. Traditional methods, leveraging on multiple low-level features, are composed of three steps: segmentation, hand-engineered feature extraction, and classification \cite{1}. Compared with this solution, deep learning methods exploit semantic, high-level, and deeper feature representations directly from data. These data-driven deep learning methods have made major breakthroughs and progress in detection accuracy and timeliness \cite{2}.

Object detection is a hot research topic for what concerns the deep learning based approaches. Due to the great variety of sensors implemented in automotive applications, deep learning methods have been developed both for 2D and 3D raw data.
Regarding methods for 2D data, the most widely used architectures take images as input: basically, they can consist of convolutional neural network (CNN) with region proposal as R-CNN \cite{3}, SPP-Net \cite{4}, Fast and Faster R-CNN \cite{5,2}, or of single-shot detectors as YOLO and SSD \cite{6,7}. However, although these methods based on camera images are widely used for object classification due to the high semantic content they provide \cite{6bis}, they are usually affected by the lack of spatial information. That leads to lower accuracy in object position estimation, and also make the process more sensitive to occlusions \cite{1}. 

Compared with 2D methods, 3D methods take point-clouds as input data. A rotating or solid-state LiDAR sensor provides 3D positions, as well as reflections intensity of each point surrounding the vehicle. This representation is more accurate and spatial-rich; thus, it can improve both the object detection and classification process and even reduce the negative effects due to occlusions. However, a 3D point-cloud generally needs to be re-arranged to be directly used as an input of a 2D object detection method. This paper deals with the problem of using a LiDAR point-cloud in a 3D object detection method to ensure object detection for an autonomous vehicle.

\begin{figure}[t!]%
\label{fig: photo}
	\centering
	\includegraphics[width=0.4\textwidth]{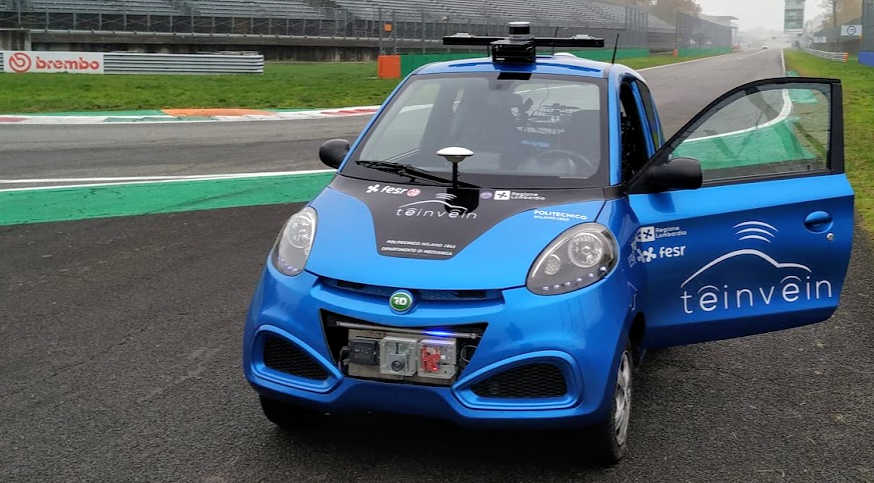}
	\caption{The experimental vehicle with LiDAR mounted on the roof}
	\label{figure:car}
\end{figure}

This paper presents a comparison between two different projection-based methods that have been implemented within the state estimation routine of a prototype of an autonomous vehicle displayed in Fig.~\ref{figure:car}. The first solution is based on the open-source Apollo FCNN-based object detection algorithm; the second is instead a geometric based pipeline for 3D point-clouds processing developed in our labs. The experimental vehicle is equipped with a 16-beams LiDAR sensor. All the experimental tests have been carried in the Monza ENI circuit~\cite{Monza}. The ground truth for the validation and comparison of the methods is based on GPS measurements of the target object corrected with RTK service. Both algorithms are implemented on ROS~\cite{quigley2009ros}, the deep learning based approach runs on an NVIDIA GTX 1050Ti, while the geometric one has been implemented to work on CPU only.

This paper is divided as follows; in Section~\ref{sec: soa} we illustrate the current state of the art concerning point-cloud based obstacle detection, with particular interest on the deep learning based approaches. The FCNN-based objects detection algorithm is presented in Section~\ref{sec: algorithm} while our geometric-based method is described in detail in Section~\ref{sec: pipeline}. The experimental setup is presented in Section~\ref{sec: vehicle} while Section~\ref{sec: results} presents the analysis of the results in terms of a comparison between the position estimates and a ground truth for both proposed solutions.

\section{Related Works}
\label{sec: soa}
State of the art point-clouds based object detection can be divided into three subcategories: projection-based methods, volumetric convolutional methods and raw point-cloud-based methods \cite{1,8}.

Projection-based methods implement a single or multi-view projection of a 3D point-cloud, resulting in a 2D grid, which is then processed to find object clusters with the desired confidence. This grid is then processed by a  2D CNN, or to a traditional pipeline. Feature extraction is done during the projection of points on a horizontal plane, discretized with an assigned grid of pre-determined dimensions. In such a way, different channels can be added to improve the number of features available (e.g., height, density, intensity, occupancy, etc.) within the corresponding grid cell. Next, these channels are stacked together and treated as 2D image. Complex-YOLO \cite{9}, BirdNet \cite{10}, PIXOR \cite{11} map point cloud into Bird's Eye View (BEV). LMNet \cite{12}, VeloFCN \cite{13} takes the frontal view (FV) of point cloud as input. MV3D \cite{14} adopts both BEV and FV of point cloud as input. BEV is widely used due to its lower probability of occlusion w.r.t FV. Projection-based methods shrink dimensions of point cloud and computational cost through projection, meanwhile causing inevitable spatial information loss. These methods actually achieve a trade-off between accuracy and computational cost. 

Volumetric convolutional methods first conduct point cloud voxelization representing 3D point cloud as regularly spaced 3D voxel grids. Features such as point intensity, height, density and occupancy are extracted manually from points within corresponding particular voxel cells. Then 3D convolutions are adopted to process these voxels \cite{15, 16, 17}. These methods encode spatial information of point cloud explicitly and have less information loss compared to projection-based methods, thus producing satisfactory detection accuracy. However, due to the expensive computational cost of 3D convolution and existing empty voxels caused by point cloud sparsity, volumetric convolutional methods are time-consuming and inefficient.

Projection-based methods and volumetric convolutional methods aim to convert point-clouds into 2D images or 3D voxel grids. Differently, raw point-cloud-based methods directly handle point-clouds to minimize spatial information loss. Most of the raw point-cloud-based methods are variants of PointNet\cite{18}, widely used for object classification and semantic segmentation. 
PointNet++ \cite{19} is the advanced version of PointNet and can extract local features efficiently, whilst Frustum PointNet \cite{20} allows to construct subsets of point-clouds based on 2D detections on the image plane. Then these subsets are fed directly into a PointNet for classification and prediction. PointNets methods assume segmented objects and are mainly used for simple and indoor scenes in robotics, so they are not widely used in autonomous driving.

In conclusion, compared to the other two subcategories, projection-based methods are well-researched in the context of driving scenarios due to its similarity to mature 2D object detection methods. Even if, based on hand-crafted feature extraction, they offer a good trade-off between time complexity and detection performance \cite{1}.



\begin{figure}[t!]%

	\centering
	\includegraphics[width=0.4\textwidth]{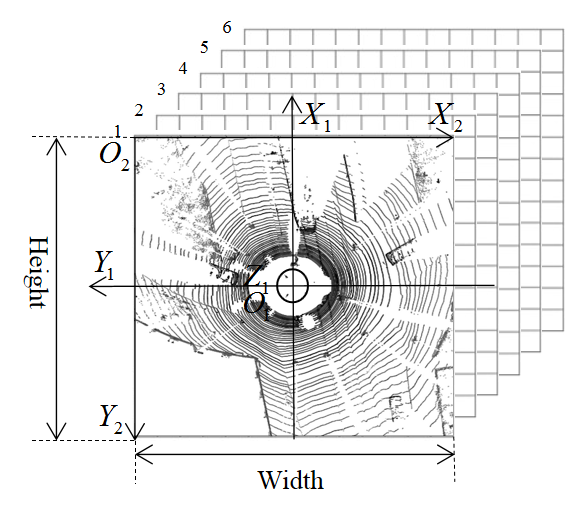}
	\caption{Bird-eye-view of point cloud with 6 channel features}
	\label{figure:channel}
\end{figure}

\begin{figure*}[t!]%

	\centering
    	\includegraphics[width=0.73\textwidth]{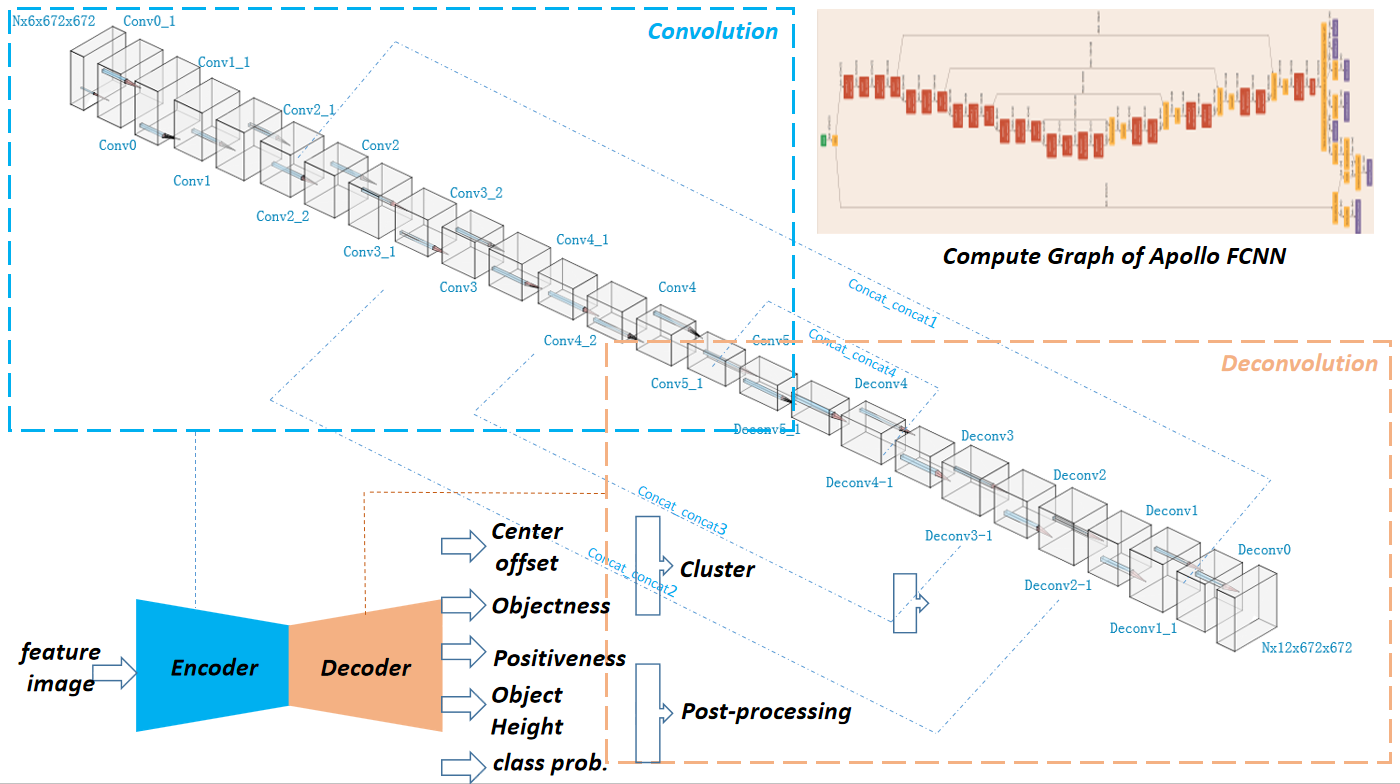}
	\caption{Detailed Architecture of Apollo FCNN-based Model}
	\label{fig:FCNN_model}
\end{figure*}

\section{Fcnn-based Object Detection}
\label{sec: algorithm}
Apollo is an open autonomous driving platform, which has released all the most important modules for the implementation of autonomous driving.
For what concerns the perception task, Apollo uses Fully Convolutional Neural Network \cite{21} (FCNN) to conduct segmentation of point-clouds provided by  LiDAR sensors. Apollo trained and tested its FCNN-based object detection model using its own large-scale dataset, ensuring high level of robustness and accuracy. 

Apollo FCNN-based object detection model is composed of 3 steps \cite{22}:
\begin{enumerate}
\item Channel feature extraction: the 3D point-cloud provided by the LiDAR is projected in a BEV image with pre-determined width, height and grid cell size. To avoid loss of information during the projection of 3D point-cloud into a 2D image, 6 additional channels shown in Fig.~\ref{figure:channel} are stacked together the new pattern to recover information about the peak and the medium values of height, intensity and distribution of the collapsed points for each cell. Moreover, binary information concerning the effective occupancy of each grid are included.
\item FCNN-based obstacle prediction: the 6 extracted feature channels, together with the BEV grid image, are provided to the Apollo FCNN detection model presented in Fig.~\ref{fig:FCNN_model}. Assuming a unitary batch size $N=1$, i.e. a unitary point-cloud frame, a 6x672x672 BEV image is used as input for the estimation process. Apollo FCNN-based model is composed of convolutional layers (named as encoder), deconvolutional layers (decoder) and prediction layers (predictor). The former part of the FCNN is defined by 15 consecutively stacked 15 convolutional layers to downsample the spatial resolution of the input BEV image and extract increasingly complicated features as the layers get deeper. Then, 10 tightly connected deconvolutional layers upsample the encoded BEV image to the spatial resolution of the input BEV image. Skip connection technique is used to promote BEV image's spatial details recovery during deconvolution process as well as better deep neural network training \cite{23}.
The algorithm provides as output a 672x672 BEV image with 6 attributes for each grid cell related to its effective occupancy, and eventually to its position, confidence prediction, class, absolute angle and height.
\item Obstacle clustering and post-processing: clustering is performed on the output BEV image accounting for the 6 attributes related to each occupied grid cell, basing on a union find algorithm. Then, during post-processing a confidence value is assigned for each candidate clustered object, in order to get a more accurate final output.
\end{enumerate}

To conclude, for each individual detected object (stationary and moving), the estimation process provides the following information:
\begin{itemize}
\item detection confidence score;
\item position w.r.t. lidar coordinate system;
\item main dimensions, i.e. length, width and height;
\item classification, e.g. car, truck, cyclist or pedestrian;
\item main distance, i.e. distance between the object's centroid and lidar origin.
\end{itemize}

ROS is used to implement and test Apollo FCNN-based 16-beams-lidar detection model. Running on a NVIDIA GTX 1050Ti, the algorithm provides estimates at 10 HZ with KITTI dataset \cite{24}. 

\section{Geometric-based Object Detection}

\label{sec: pipeline}
\begin{figure}[t!]%
	\centering
	\includegraphics[width=0.35\textwidth]{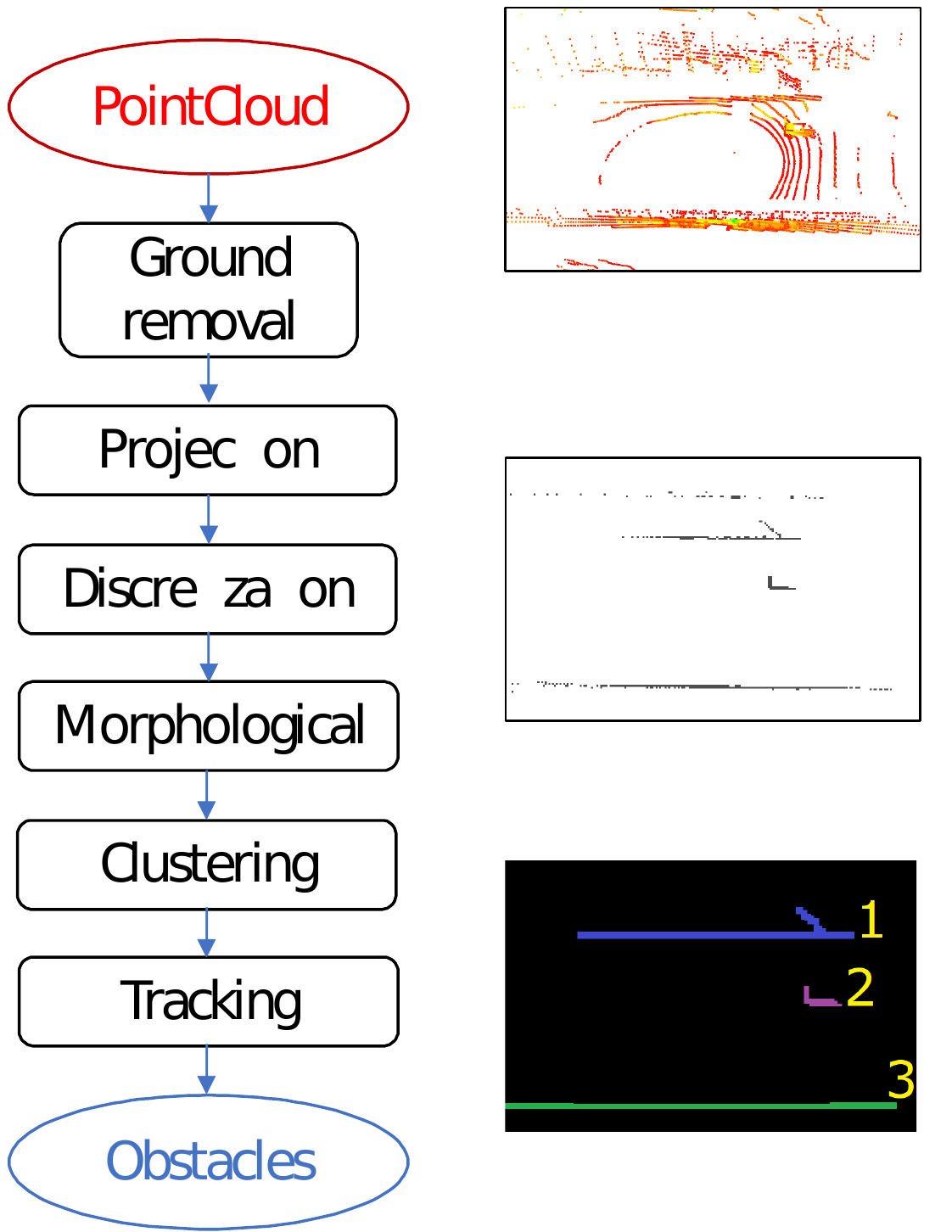}
	\caption{Representation of the different steps of the point-cloud elaboration pipeline. On the right are displayed three images of the main steps of the process, the input image, the discretized 2D grid and the tracked obstacles.}
	\label{fig:pipeline}
\end{figure}

As stated in the introduction, the second analyzed approach does not leverage on deep learning but only on geometric and morphological transformation to retrieve the obstacles from the 3D point-cloud.
The pipeline in Fig.~\ref{fig:pipeline} show the operations required to convert a set of 3D points into a list of obstacles on the horizontal plane. The presented pipeline handles the conversion from a 3D point-cloud to a 2D occupancy grid, including the final tasks of clustering, identification and tracking.

The first step of this pipeline is to remove all the detected points belonging to the ground plane, i.e. the road surface, in order to reduce the incidence of false positives in the estimation process. To perform this task, an approach similar to the one presented in \cite{zermas2017fast} is implemented, in which the plane fitting problem is based on RANSAC (RANdom SAmple Consensus). 

Once the ground plane is removed, all the remaining points are most likely  belonging to obstacles. Thus, they are projected on a 2D plane 
to obtain a set of 2D points on a plane parallel to the road surface. 
Discretization is then carried out through the application of a grid on the identified horizontal plane. In particular, the grid is divided in square cells with side equal to $0.3 \, m$: if the number of points in the cell is higher than a pre-computed threshold, the cell is set to occupied. 
Then, a further threshold parameter is applied to filter out noise effects that may eventually lead to false positives. Both the described parameters have been tuned basing on experimental measurements in controlled environment, considering decreasing values depending on the radial distance from the sensor to take into account the variable density of the point-cloud~\cite{mentasti2019multi}. 

The output of the previous phase is a simplified representation of the area surrounding the ego-vehicle, which is used to obtain the relative position of each object close to the vehicle.
The occupancy grid provides information regarding the presence of objects for each cell, hence clustering is required to merge elements in the  occupancy grid. As preliminary step, a set of morphological operations (i.e., opening followed by closure) is required to connect areas that might belong to the same object but are not adjacent. This might happen due to obstructions or to the particular shape of the object itself, that caused the number of points belonging to a specific cell to be lower than the filtering threshold explained before. 
The result is still an occupancy grid where all elements belonging to an obstacle are connected. 

Clustering is based on the OpenCV \cite{opencv_library} implementation of SAUF (Scan plus Arraybased Union-Find) \cite{wu2009optimizing}; the output is a list of all the connected components in the occupancy grid which belong to real obstacles, defined by relative position of the respective centre of symmetry (CoS) with respect to the ego-vehicle and its equivalent dimensions $\rho_{o_{i}}$. Length and width of each identified object are provided, but the accuracy of those data is limited due to the nature of the input, an obstacle directly in front of the sensor will be visible only on its back, and therefore it will not be possible to  compute its length. Thanks to numerous optimization, and the filtering process performed at the beginning of the pipeline this solution is able to run smoothly at $20\, Hz$ (i.e., the maximum frequency of the LiDAR sensor) on a consumer laptop.

\section{System setup}
\label{sec: vehicle}

The developed prototype of autonomous vehicle is shown in Fig.~\ref{figure:car}, a front wheel drive full-electric plug-in quadricycle, powered by a 6 kW motor. Starting from a commercial vehicle, the modifications made on the most important actuation systems (i.e. the steering, throttle and braking systems, as presented in \cite{Vignati, Bersani}) allowed to obtain a vehicle fully automated. Therefore, in the current configuration the vehicle takes references from a control unit on which a trajectory planner like the one presented in \cite{traba} runs in real-time. Moreover, the control loop is closed by the estimation algorithm presented in \cite{Stima2019}, which provides vehicle pose and speed estimations at $20\, Hz$ running on a consumer laptop.

A second vehicle is required to test and compare the presented algorithms. Thus, this vehicle represents the target of the estimation process, where the ground truth is given by a GPS mounted on its roof to measure its absolute position with RTK correction service. Since the LiDAR sensor is mounted in correspondence of the ego-vehicle center of gravity, a transformation of coordinates involving the current heading angle of the ego-vehicle is required to have the ground-truth in the same local reference system of the LiDAR, i.e. the same in which estimates of the obstacle vehicle are provided. 
The transformation of coordinates is presented in eq. \eqref{eq: trans}, where the output vector $z_{loc}$ measures the current longitudinal and lateral distance ($x_{loc}$ and $y_{loc}$ respectively) between the vehicle and the obstacle in the local reference frame, i.e. the right-handed moving reference system of the vehicle. This can be obtained by rotating the vector $z_{abs}$, which contains the distance between vehicle and obstacle in the absolute reference frame, obtained by converting the GPS measurements from degrees to meters in UTM coordinates.

\begin{equation}
\label{eq: trans}
z_{loc} = 
     R \cdot z_{abs} = \begin{bmatrix} cos(\psi)\quad sin(\psi)\\ -sin(\psi) \quad cos(\psi)\end{bmatrix}  \begin{bmatrix} \Delta X \\ \Delta Y\end{bmatrix}_{abs}
\end{equation}

To conclude, the rotation matrix $R$ allows a clockwise rotation accounting for the current angle between the absolute and the local reference frame, i.e. the heading angle of the vehicle ($\psi$).

\begin{figure}[t!]%
	\centering
	\includegraphics[width=0.45\textwidth]{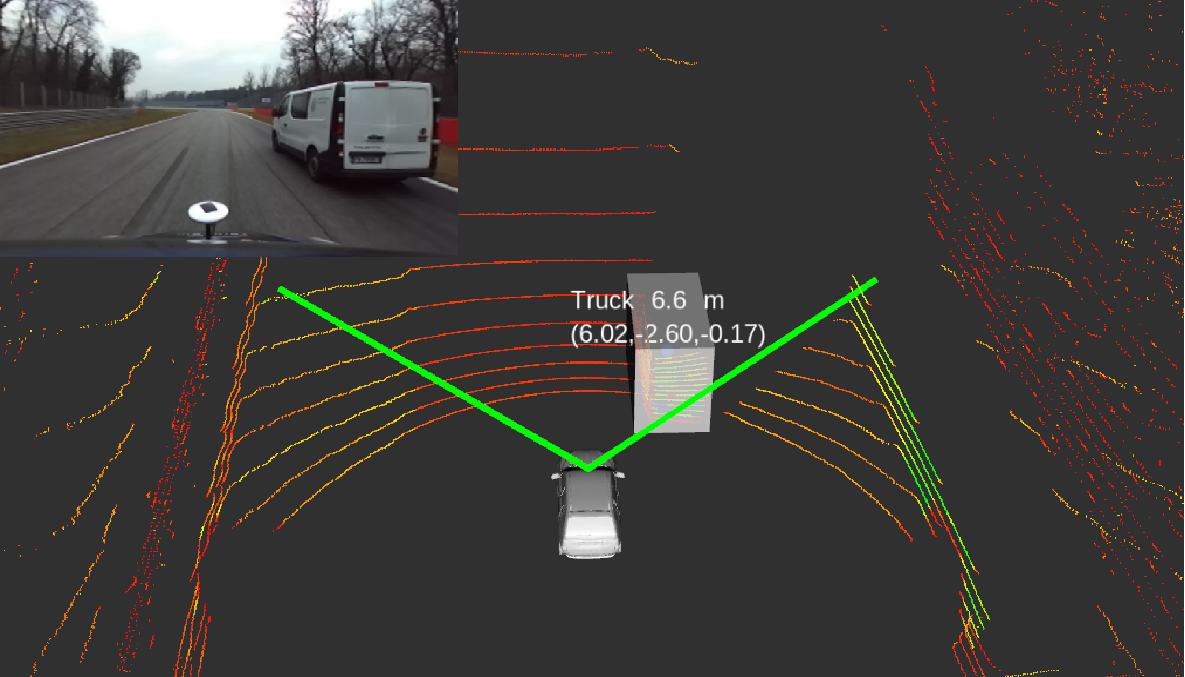}
	\caption{Dataset visualization}
	\label{fig: data}
\end{figure}

\section{Experimental Results}
\label{sec: results}
To quantitatively compare the two algorithms we recorded a dataset at the Monza ENI Circuit, using the setup described in the previous section.
In particular, the dataset refers to the part of the track between the \emph{Serraglio} and the \emph{Parabolica}. This allows to evaluate the performances in a mixed section of the circuit, i.e. where the relative direction between the vehicle and the target obstacle changes frequently. Moreover, the dataset includes two overcoming manoeuvres, hence the two vehicles are changing their respective positions.
The algorithms' performances are evaluated in terms of estimated distance in longitudinal and lateral directions in the vehicle reference frame, and also in terms of predicted sizes of the target vehicle, a commercial van whose width and length are respectively equal to $2$ and $5$ meters. A visual representation of the system's output is given in Fig.~\ref{fig: data}, where the bounding-box created by the Apollo FCNN is superimposed to the target vehicle that is caught on camera as shown in the top-left part of the image.

\begin{figure}[t!]%
	\centering
	\includegraphics[trim=20 0 0 10,clip,width=0.52\textwidth]{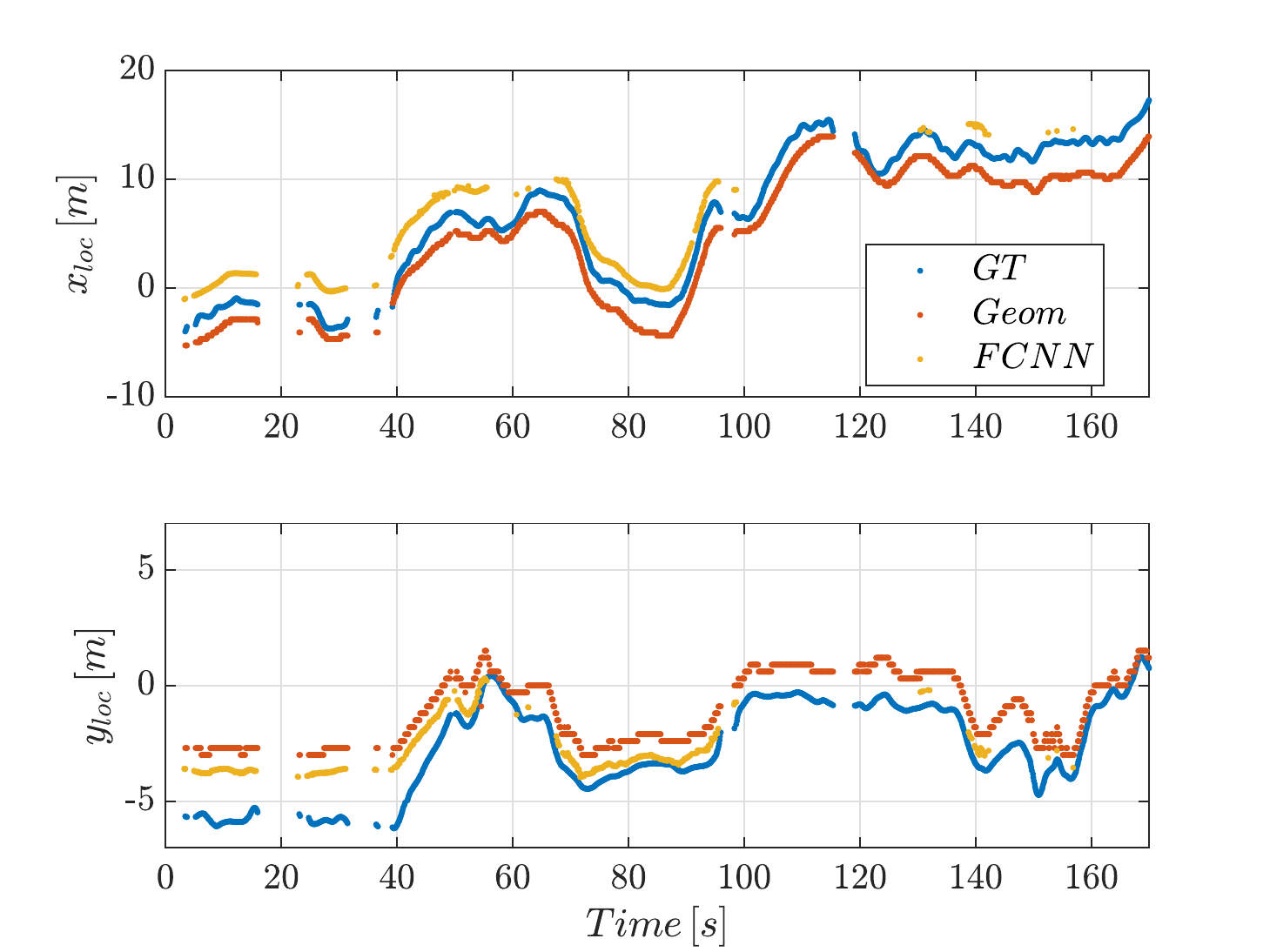}
	\caption{Comparison of estimates with the ground truth (GT)}
	\label{fig: stime}
\end{figure}

To compare the results we run both algorithms on the recorded data and compared the computed positions against the ground-truth. A visual comparison of the estimated distances between vehicle and obstacle (i.e., $x_{loc}$ in longitudinal and $y_{loc}$ in lateral direction), is shown in Figure~\ref{fig: stime}. It is possible to notice how the measurement direction impacts on the algorithm performance. Although both algorithms qualitatively follow the ground truth correctly, the geometrical one underestimates the distance $x_{loc}$ with an offset with average value $\delta_{G} = -2.09 \,m$ and standard deviation $\sigma_{G} = 1.25 \,m$; on the other hand, the estimation performed with FCNN points out positive offset with average $\delta_{F} = 1.77 \, m$ with a smaller standard deviation $\sigma_{F} = 0.54 \, m$ compared to the previous one. For what concerns the different signs of the offsets, the geometrical method provides the distance with respect to the centroid of the 3D points measured by the sensor, hence the fact that the obstacle is higher than the vehicle affects the accuracy of the method, since it cannot account for the real shape of the target. On the other hand, the FCNN used by the other algorithm has been trained with point-clouds obtained from L-shaped vehicles, that explains also why estimates are not available when the ground truth of lateral distance $y_{loc}$ is close to zero, i.e. when target and the vehicle are driving in same direction (as shown in the range $Time \in [100, 130] s$). Thus, the positive value of $\delta_{F}$ is due to the fact that the GPS receiver on the target is mounted $1.2\,m $ ahead of the back, hence is located $1.3 \, m$ behind the real centroid of the vehicle: this explains why the FCNN gives a higher value of $x_{loc}$. 

About the estimated distance in lateral direction, results in the bottom of Fig.~\ref{fig: stime} show in general a larger relative error, i.e. the value of the offset compared to the nominal value given by the ground truth. Moreover, as anticipated, the method based on FCNN is not able to provide any obstacle detection in vehicle-following scenarios. Furthermore, as for the estimates of $x_{loc}$, the fact that the GPS receiver is located $1$ meter beyond the lateral edge of the obstacle generates an offset whose sign depends directly on the direction of the lateral displacement between the two vehicles.
Moreover the geometric solution seems to perform better at high distance (i.e., above $10m$), and when the obstacle is directly in front of the vehicle, where the number of points representing the van is considerably low and the neural network is not able to extract enough features to detect it. Conversely the deep learning based approach performs considerably better on the lateral position in close distance, where the number of points is high and the network can accurately reconstruct the bounding box. 


To conclude, the estimated width and length of the vehicle for each algorithm are shown in Table~\ref{tab: w&l}. Those values are derived together with the estimates shown in Fig.~\ref{fig: stime}. Even for this comparison, it is possible to state how the results are similar, taking into account that the lower standard deviations obtained by the FCNN-based algorithm can be ascribed to the fact that it does not provide any prediction in scenarios far from the optimal working conditions.


\section{Conclusions}

This paper presents a comparison between two algorithms for obstacles state estimation in autonomous driving. The analyzed algorithms take as input 3D point-clouds given by a rotating LiDAR sensor, to estimate the components in lateral and longitudinal directions of the distance between the vehicle on which the sensor is mounted and the target one. Moreover, both algorithms give as output an estimate of the main obstacle dimensions for each detection. 

In this paper 3D point-clouds are pre-processed by performing a projection on a 2D plane, that is firstly discretized and then processed, using geometrical and morphological analysis in the first case, or is fed to a FCNN in the other one.


Results point out that the estimates provided by the algorithm based on FCNN and deep learning are in general less affected by noise, but this algorithm does not work properly when the two vehicles are one in front of the other and they are moving in the same direction, i.e. when LiDAR can detect only one edge of the target vehicle. This can be due to the dataset used to train the network. However, this method is more accurate when dealing with the estimation of the main sizes of the vehicle. On the other hand, the presented paper shows how the algorithm that performs geometrical and morphological analysis on the pre-processed point-cloud is more flexible, since it provides estimates in a wider range of working conditions. 

\begin{table}
\caption{Estimates of obstacle main dimensions}
\label{tab: w&l}
\centering
\begin{tabular}{lrr}
\hline
  & Geom & FCNN\\ \hline
  \rule{0pt}{2ex}    
$E[l]\, [m]$ & 4.03 & 4.48\\
$\sigma_{l}\, [m]$ & 1.7 & 0.6\\ 
$E[w]\, [m]$ & 2.12 & 2.25\\
$\sigma_{w}\, [m]$ & 0.37 & 0.23\\

\hline
\end{tabular}
\end{table}

\bibliographystyle{./bibliography/IEEEtran}
\bibliography{./bibliography/IEEEabrv,./bibliography/IEEEexample}

\end{document}